\documentclass[review,11pt]{ReportTemplate}
\usepackage{bm}
\usepackage{graphicx}
\usepackage{paralist}
\usepackage{amsmath,mathrsfs,amssymb}
\usepackage{booktabs}
\usepackage{multirow}
\usepackage[ruled,linesnumbered]{algorithm2e}
\usepackage{capt-of}

\begin{document}
\begin{frontmatter}
\title{Soft Gradient Boosting Machine}

\author{Ji Feng$^{1,2}$}
\author{Yi-Xuan Xu$^{1,3}$}
\author{Yuan Jiang$^{3}$}
\author{Zhi-Hua Zhou$^{3}$\\ \texttt{fengji@chuangxin.com, \{xuyx, jiangy, zhouzh\}@lamda.nju.edu.cn}}
\address{$^1$Sinovation Ventures AI Institute \\ $^2$Baiont Technology \\ $^3$National Key Laboratory for Novel Software Technology, Nanjing University, Nanjing 210093, China}

\begin{abstract} 
Gradient Boosting Machine has proven to be one successful function approximator and has been widely used in a variety of areas. However, since the training procedure of each base learner has to take the sequential order, it is infeasible to parallelize the training process among base learners for speed-up. In addition, under online or incremental learning settings, GBMs achieved sub-optimal performance due to the fact that the previously trained base learners can not adapt with the environment once trained. In this work, we propose the soft Gradient Boosting Machine (sGBM) by wiring multiple differentiable base learners together, by injecting both local and global objectives inspired from gradient boosting, all base learners can then be jointly optimized with linear speed-up. When using differentiable soft decision trees as base learner, such device can be regarded as an alternative version of the (hard) gradient boosting decision trees with extra benefits. Experimental results showed that, sGBM enjoys much higher time efficiency with better accuracy, given the same base learner in both on-line and off-line settings.
\end{abstract} 
\end{frontmatter}

\section{Introduction}
Gradient Boosting Machine (GBM) \cite{friedman2001greedy} has proven to be one successful function approximator and has been widely used in a variety of areas \cite{bennett2007netflix,chapelle2011yahoo}. The basic idea is to train a series of base learners that minimize some predefined differentiable loss function in a sequential fashion. When building such learning devices, non-differentiable decision tree~\cite{quinlan1993c4} are often used as base learner. For instance, Gradient Boosting Decision Tree (GBDT) \cite{friedman2001greedy} and its variant implementations such as XGBoost \cite{chen2016xgboost}, LightGBM \cite{ke2017lightgbm}, and CatBoost \cite{prokhorenkova2018catboost} are one of most widely used versions. Such models are still the best choice for tabular data and successful applications raging from collaborative filtering \cite{bennett2007netflix} to information retrieval \cite{chapelle2011yahoo}, and to particles discovery \cite{baldi2014searching}. However, it is still an open problem for GBM models to be used when facing streaming data since the base models could not adapt to the environment once trained.

Differentiable programming, on the other hand, requires not only the loss function to be differentiable, but also the learning modules. Concretely, by constructing several differntiable learning modules into any DAG (Directed Acyclic Graph) form, the whole structure can be jointly optimized via stochastic gradient descent or its variant optimization methods. Such system have several appealing properties including representation learning, scalability and can be used in an online fashion.

The work of using multi-layered gradient boosting decision trees (mGBDT) \cite{feng2018multi} for representation learning is one seminal work trying to combine the best part of both worlds. Concretely, mGBDT can have hierarchical representation learning ability as any differentiable programming models do, but still keeps the non-differentiable property so as to handle tabular data in a better way. This work opens many new opportunities and challenges and there are still much to explore. 

Inspired by mGBDT, in this work, we flipped the challenge around and ask, instead of building a GBM behaves like a differentiable program, \textit{``Can we build up a differentiable system behaves like a non-differentiable gradient boosting machine?"} To achieve this, we propose to build a soft version of gradient boosting machine by concatenating several differentiable base learners together, and by introducing local and global loss inspired from GBMs, the whole structure can be jointly optimized. In addition, we propose to use soft decision trees \cite{irsoy2012soft,frosst2017distilling} as base learner, and the corresponding soft GBDT can be regarded as an alternative choice for XGBoost \cite{chen2016xgboost} when hard decision tree is not the best fit. There are several advantages of designing such a device as follows: 

Firstly, compared with traditional (hard) gradient boosting machine, the soft gradient boosting machine is much faster to train. Instead of training base learners one at a time, soft GBM is able to simultaneously train all base learners. Empirical results showed that, given the same base leaner, experiments on several benchmark data sets can give over 10x times of speed up, with even better accuracy. In addition, when fitting a traditional GBM model, one base learner has to ``see'' all the training data before moving to the next learner, making the system not suitable for incremental or online learning, whereas the soft GBM is bornt with the ability under such settings. 

Secondly, current implementations for GBDT such as XGBoost used CART \cite{breiman1984classification} as base learners, making it less straightforward when facing multi-dimensional regression tasks. sGBDT, on the other hand, can naturally handle such tasks with soft trees as base learner. Such property also makes sGBDT more suitable for knowledge distillation or twice learning \cite{zhou2004nec4,hinton2015distilling}, since the distillation process with transform the classification one hot labels into a dense vector on the training set. 

Finally, due to local and global loss injections, soft GBM gives an exponential increase in terms of the interactions between base learners, making the system more efficient and effective compared with soft averaging several base learners. Although beyond the scope of this paper, this give some more room for theoretical analysis for sGBMs compared with other soft ensemble methods.

The rest of the paper is organized as follows: first, some related works are discussed; second, details on our proposed method are presented; finally, empirical experiments and conclusions are illustrated and discussed.

\section{Related Work}

Decision tree \cite{quinlan1993c4} are powerful model with excellent interpretability and best used in modeling tabular data. Tree models are usually used in an ensemble learning fashion \cite{zhou2012ensemble}. For instance, decision tree boosting\cite{freund1997decision} is one of the most used ensemble techniques in a variety of areas \cite{viola2001robust,schapire2000boostexter} with profound theoretical grantees. Gradient Boosting Trees \cite{friedman2001greedy} tries to build the boosting procedure as an optimization task over a differentiable loss function in an iterative fashion. Its variant implementations such as XGboost \cite{chen2016xgboost}, LightGBM \cite{ke2017lightgbm}, and CatBoost \cite{prokhorenkova2018catboost} are still the dominant models when facing tabular or desecrate typed data. 

There has been some seminal works in combining neural networks with tree structures or ensemble learning in general. The method of mixture of experts \cite{jordan1994hierarchical}, for instance, used a tree-like routing structure to train several neural networks via the EM algorithm. Other approaches include building differentiable tree ensembles on top of a neural network \cite{kontschieder2015deep}, or using ensemble methods on neural networks \cite{martinez2019sequential,shalev2014selfieboost,hehn2019end,zhou2002ensembling}.

Knowledge distillation \cite{hinton2015distilling} or twice learning \cite{zhou2004nec4} is an attempt to squeeze knowledge from a big and complex model into a small and simple one. The basic idea is to train the small and simple model to mimic the behavior of the big and complex model by producing a new training set with twice labelling\cite{frosst2017distilling,zhou2004nec4},such technique is not only useful for model compression, but also of great interest when the smaller model can be easily interpreted.  

Recently, there has been works on realizing deep learning models via non-differentiable modules. For instance, the work of deep forest \cite{zhou2017deep} is the first work trying to build a non-differentiable system while still enjoys the benefit of deep models. The mGBDT model \cite{feng2018multi} is the first attempt to achieve representation learning via an multi-layered gradient boosting decision trees. In this work, we flipped the challenge around and ask, can we build a differentialbe system, while keeping all the benefit of it, and enjoys some benefit from the non-differentiable world such as excellent tabular processing ability such including XGBoost? In the following section, a detailed description of the proposed method is presented.

\section{The Proposed Method}
Before presenting details on the proposed method, we first give a very brief introduction on Gradient Boosting Machine (GBM) to make this paper self-contained. Concretely, given a training dataset $\{\mathbf{x}^i,y^i\}_{i=1}^N$, the goal of GBM is to obtain a good approximation of the function $F^{*}(\mathbf{x})$ that minimizes the empirical loss $\sum_{i=1}^N l (F(\mathbf{x}^i), y^i)$.\ GBM assumes that $F^{*}(\mathbf{x})$ has the additive expansion form: $F(\mathbf{x})=\sum_{m=0}^M \beta_m h_m(\mathbf{x};\boldsymbol{\theta}_m)$, 
where $h_m(\mathbf{x};\boldsymbol{\theta}_m)$ is parametrized by $\boldsymbol{\theta}_m$, and $\beta_m$ is the coefficient of $m$-th base learner.

\begin{figure}[t]
    \centering
    \includegraphics[width=\columnwidth]{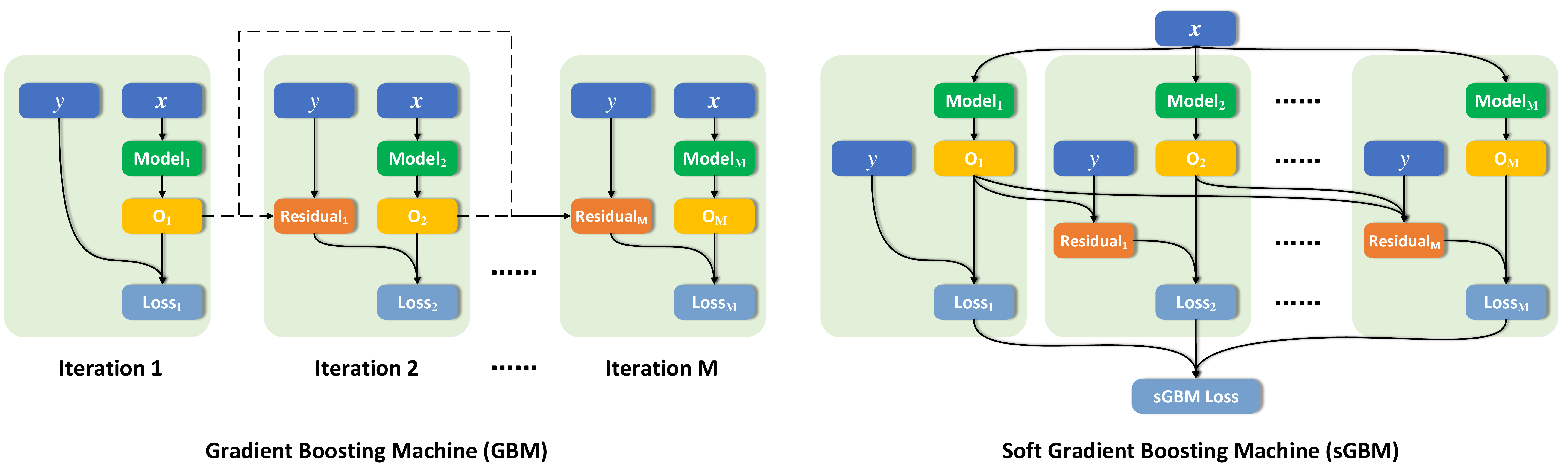}
    \caption{Graphical illustration of GBM and sGBM}
\label{illustration}
\end{figure}

The training procedure of GBM is then to learn parameters $\{h_m(\mathbf{x};\boldsymbol{\theta}_m), \beta_m\}_{m=0}^{M}$ from training data. GBM first assumes that $\beta_0 h_0(\mathbf{x};\boldsymbol{\theta}_0) \equiv 0$, then, $h_m(\mathbf{x};\boldsymbol{\theta}_m)$ and $\beta_m$ are determined in a sequential fashion as follows: First, given $y^i$ and the prediction of GBM: $F_{m-1}(\mathbf{x})=\sum_{j=0}^{m-1}\beta_j h_j(\mathbf{x};\boldsymbol{\theta}_j)$
obtained from the previous round, GBM computes the so-called residual for each training sample: $r_{m}^i = - \frac{\partial l(F_{m-1}(\mathbf{x}^i), y^i)}{\partial F_{m-1}(\mathbf{x}^i)}$.
Second, the next base learner $h_m$ is fitted towards the residuals.\ Coefficient $\beta_m$ is then determined by either least squares or a constant \cite{chen2016xgboost,ke2017lightgbm}. Finally, once we have updated parameters of learner $h_m$ and coefficient $\beta_m$, we can use it as given and update the prediction of GBM: $F_m(\mathbf{x}) = F_{m-1}(\mathbf{x}) + \beta_m h_m(\mathbf{x};\boldsymbol{\theta}_m)$.

Then the training procedure can move on to the next round. The whole training procedure is summarized in Algorithm \ref{train_GBM} and illustrated in the left part of Figure \ref{illustration}. Following the convention \cite{chen2016xgboost,haihao2019accelerating}, we replace all learner coefficients with one fixed coefficient $\epsilon$.

\begin{algorithm}[h]
	\label{train_GBM}
	\caption{Training (hard) GBM}
		\KwIn{Training data $\{\mathbf{x}^i,y^i\}_{i=1}^N$, number of base learner $M$, learner coefficient $\epsilon$}
		\KwOut{Trained GBM $F_M(\mathbf{x})$}
		$F_0(\mathbf{x}) \leftarrow 0$ \tcp*{Initialize}
		\For{$m=1\ to\ M$}{
		$r_{m}^i \leftarrow - \frac{\partial l(F_{m-1}(\mathbf{x}^i), y^i)}{\partial F_{m-1}(\mathbf{x}^i)}$\quad for $i=1,\cdots,N$ \tcp*{Residual}
		$\boldsymbol{\theta}_m \leftarrow \mathop{\arg\min}_{\boldsymbol{\theta}} \sum_{i=1}^N \left \| r_{m}^i-h_m(\mathbf{x}^i;\boldsymbol{\theta}) \right \|_2^2$ \tcp*{Fit one base learner}
		$F_m(\mathbf{x}) \leftarrow F_{m-1}(\mathbf{x}) + \epsilon h_m(\mathbf{x};\boldsymbol{\theta}_m)$ \tcp*{Update GBM}
		}
		\Return $F_M(\mathbf{x})$ \;
\end{algorithm}

From the description above, it can be shown that it is hard to parallelize the training procedure for GBMs since one base learner has to be fitted before moving on to the next one. In addition, such algorithm is hard to be applied in an online fashion for the same reason.

To solve the above issue, here we introduce the soft gradient boosting machines (sGBM) by first assuming all the base learners to be differentiable. Then, instead of a soft average for concatenation of base learners, we propose to use two types of loss functions, one locally and one globally, and injecting both into the training procedure so as to make base learners to have exponential interactions and achieve the gradient boosting effect (rather than a soft average of all base learners).

\begin{algorithm}[htbp]
	\label{train_sGBM}
	\caption{Training sGBM}
	\KwIn{Training batches $\mathcal{B}=\{B_1, B_2, \cdots, B_{|\mathcal{B}|} \}$, number of base learner $M$, current sGBM parameters $\boldsymbol{\theta}=\{\boldsymbol{\theta}_m\}_{m=1}^M$}
	\KwOut{Updated sGBM parameters $\boldsymbol{\theta}$}
	\For{$b=1\ to\ |\mathcal{B}|$}{
		$o_0^i \leftarrow 0$\quad for $\mathbf{x}^i \in B_b$ \tcp*{Initialize}
		\For{$m=1\ to\ M$}{
			$o_m^i \leftarrow h_m(\mathbf{x}^i;\boldsymbol{\theta}_m)$\quad for $\mathbf{x}^i \in B_b$ \tcp*{Data forward}
			$r_m^i \leftarrow - \frac{\partial l(\sum_{j=0}^{m-1} o_j^i, y^i)}{\partial \sum_{j=0}^{m-1} o_j^i}$\quad for $\mathbf{x}^i \in B_b$ \tcp*{Residual}
			$l_m \leftarrow \sum_{\mathbf{x}^i \in B_b} \left \| r_m^i - o_m^i \right \|_2^2$ \tcp*{Local learner loss}
		}
		$\mathcal{L} \leftarrow \sum_{i=1}^M l_m$ \tcp*{Global sGBM loss}
		Update $\boldsymbol{\theta}$ w.r.t $\mathcal{L}$ using gradient descent \tcp*{Update sGBM}
	}
	\Return $\boldsymbol{\theta}$ \;
\end{algorithm}

Concretely, denote $M$ to be the number of differentiable base learners, each parametrized by $\boldsymbol{\theta}_m$. Here $M$ is a predefined number specifying how many base learner to use prior to training. Just like hard GBMs, the output for sGBM $O^i$ is the summation of all the outputs of base learners: $O^i=\sum_{m=1}^M o_m(\mathbf{x}^i,\boldsymbol{\theta}_m)$. During training, a final loss for the whole structure is defined as: $\mathcal{L}=\sum_{m=1}^M l_m$, where $l_m$ is the loss for base learners, which in turn can be further defined as $l_m = \left \| r_m - o_m \right \|_2^2$, where $o_m$ is the output of current learner $h_m$, and $r_m$ is the corresponding residual $r_{m} = - \frac{\partial l(\sum_{j=0}^{m-1} o_j, y)}{\partial \sum_{j=0}^{m-1} o_j}$.  

The right part of Figure \ref{illustration} gives a graphical illustration of the proposed structure. Since the flow of input data forms a loop-free DAG (Directed Acyclic Graph), the whole structure can be trained via SGD or its variant by minimizing both local and global loss objectives, as illustrated in Algorithm \ref{train_sGBM}.

\section{Soft Gradient Boosting Decision Tree}
Previous section gave a general introduction of the proposed sGBM without specifying which base learner to use. In this section, we give a concrete example when the base learner belongs to the decision tree family.

As one of the most applied instance of GBM, Gradient Boosting Decision Tree (GBDT) uses hard (and usually shallow) binary decision trees as base learner. Specifically, each non-leaf node inside the hard decision tree forms an axis-parallel decision plane, and each input samples will be routed to either left or right child node according to the corresponding decision plane. Such procedure is recursively defined until the input data reached the leaf node. The final prediction is the class distribution inside the leaf node which the input sample reside in.

Successful implementations of GBDTs such as XGboost \cite{chen2016xgboost}, LightGBM \cite{ke2017lightgbm}, and CatBoost \cite{prokhorenkova2018catboost} have proven to be one of the best data modeling tools especially for tabular data.  

\begin{figure*}[b]
    \centering
    \includegraphics[width=\columnwidth]{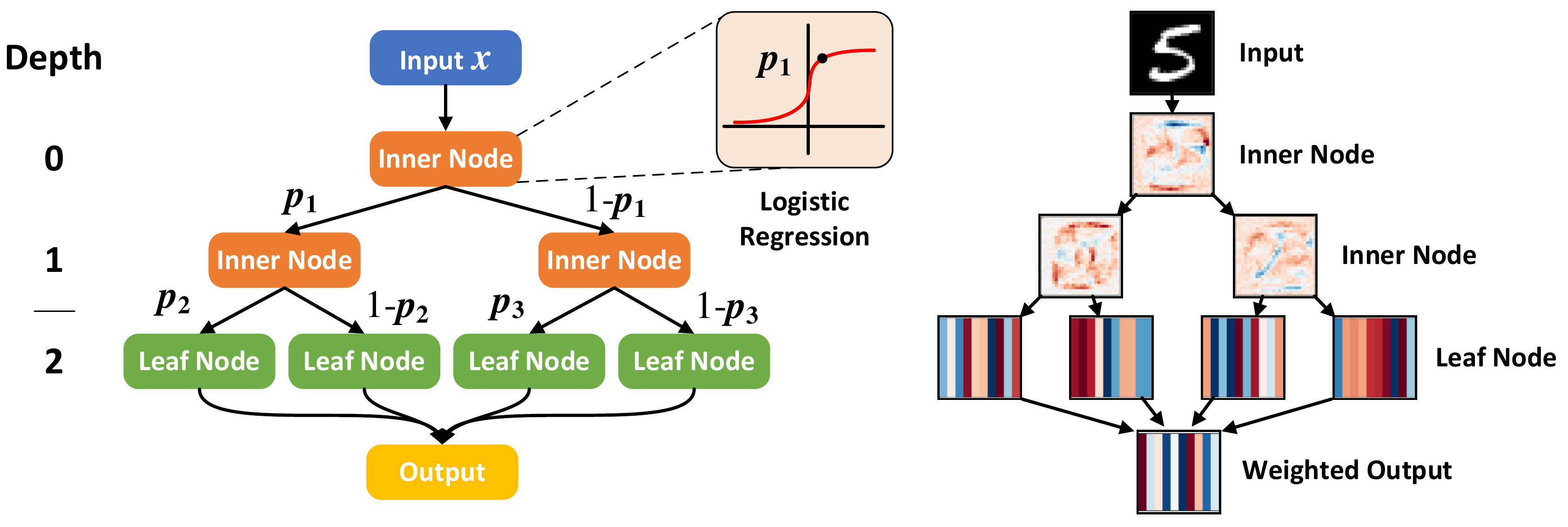}
    \caption{Graphical illustration of a single soft decision tree}
\label{fig:sdt}
\end{figure*}

Soft decision trees, on the other hand, use logistic units as the routing gate withing the internal non-leaf nodes, and the final prediction for the input sample is the weighted sum of class distributions among \textit{all} leaf nodes, where the weight is determined by the logit products on internal nodes along the decision paths. Such structure can be trained via stochastic gradient descent, a graphical illustration can be found in Figure \ref{fig:sdt}. 

When using soft decision trees as base learner, the corresponding soft GBDT has several advantage over hard GBDTs. First, hard GBDT is not the best choice when facing streaming data for apparent reasons. However, since sGBDT is parametrized and differentiable, the whole system can be fine tuned to adapt the environment more quickly. Second, when facing multi-output regression tasks, hard GBDT will have to make individual dimension per tree, making it less efficient during training. Finally, soft GBDT is much faster to train since now all the trees can be trained simultaneously.

\section{Experiment}

In this section, we designed several experiments aiming to validate the effectiveness of using soft GBMs. Concretely, given the \textit{same} base learner, we wish to compare and contrast the performance of soft GBMs over hard GBMs in terms of 1) accuracy gain; 2) training time reduction; 3) multi-output regression; 4) under incremental learning setting and 5) knowledge distillation ability.

The data sets we used includes discrete or tabular typed data such as UCI-Yeast, UCI-Letter and UCI-USPS \cite{lichman2013uci}. We also used several benchmark image datasets such as MNIST \cite{lecun1998gradient}, Fashion-MNIST \cite{xiao2017online} and CIFAR-10 \cite{krizhevsky2009learning}.

For differentiable base learners, we used multi-layered perceptrons (MLP), Convolutional Neural Networks (CNN) and soft decision trees. For simplicity, we denote sGBM$_{model}$ as one soft GBM with $model$ as its base learners.The training procedure for all sGBMs is Adam \cite{kingma2014adam}, with batch size of 128.

\subsection{Performance Comparison}
In this section, we trained several sGBMs with different base learners and compared the performance with its corresponding hard GBMs. We also reported performance of soft Averaging, whereas the voting weights can be learned in an end to end fashion. For base learners, we used MLP, CNN and decision trees, and set the number of base learners to be 10 for all soft and hard GBMs.  

We used the same soft GBDT structure across all datasets, that is, we set the number of soft trees to be 10 and tree depth to be 5. For a fair comparison, we used XGBoost having the same number of trees and tree depth. Since it is practically impossible to use a single MLP architecture across different data sets, in the experiments, we used the best MLP structure that can be found in previous literature as in \cite{feng2018multi,zhou2017deep}, details can be found in Table 1. For CNN structure, we used two consecutive convolutional layers having ReLU activations with max pooling in between, each convolutional layer has 6 and 16 feature maps with 5 by 5 kernel size, respectively. Dense layers of size $(120-84-10)$ were appended accordingly.  

We also implemented an alternative way of soft ensemble, namely soft averaging, by concatenating all the differential base learners with a linear weight. Once the final classification loss is defined at top level, all base learners can be trained simultaneously. This soft ensemble scheme can be used as a benchmark against our soft ensemble model.

\begin{minipage}[b]{0.5\linewidth}
\centering
\captionof{table}{MLP architecture over different datasets}
\vspace{0.4cm}
\begin{tabular}[b]{rr}
    \toprule
		\multicolumn{1}{r|}{Dataset} & Model Structure   \\ \midrule
		\multicolumn{1}{r|}{\emph{Yeast}}         & $\text{Input}-50-30-\text{Output}$  \\
		\multicolumn{1}{r|}{\emph{Letter}}        & $\text{Input}-70-50-\text{Output}$ \\
		\multicolumn{1}{r|}{\emph{USPS}}          & $\text{Input}-512-512-\text{Output}$ \\
		\multicolumn{1}{r|}{\emph{MNIST}}         & $\text{Input}-512-512-\text{Output}$ \\
		\multicolumn{1}{r|}{\emph{Fashion-MNIST}} & $\text{Input}-512-512-\text{Output}$ \\
		\multicolumn{1}{r|}{\emph{CIFAR-10}}      & $\text{Input}-512-512-\text{Output}$ \\
	\bottomrule
\end{tabular}
\end{minipage}\hfill
\begin{minipage}[b]{0.45\linewidth}
    \centering
	\includegraphics[width=2.75in]{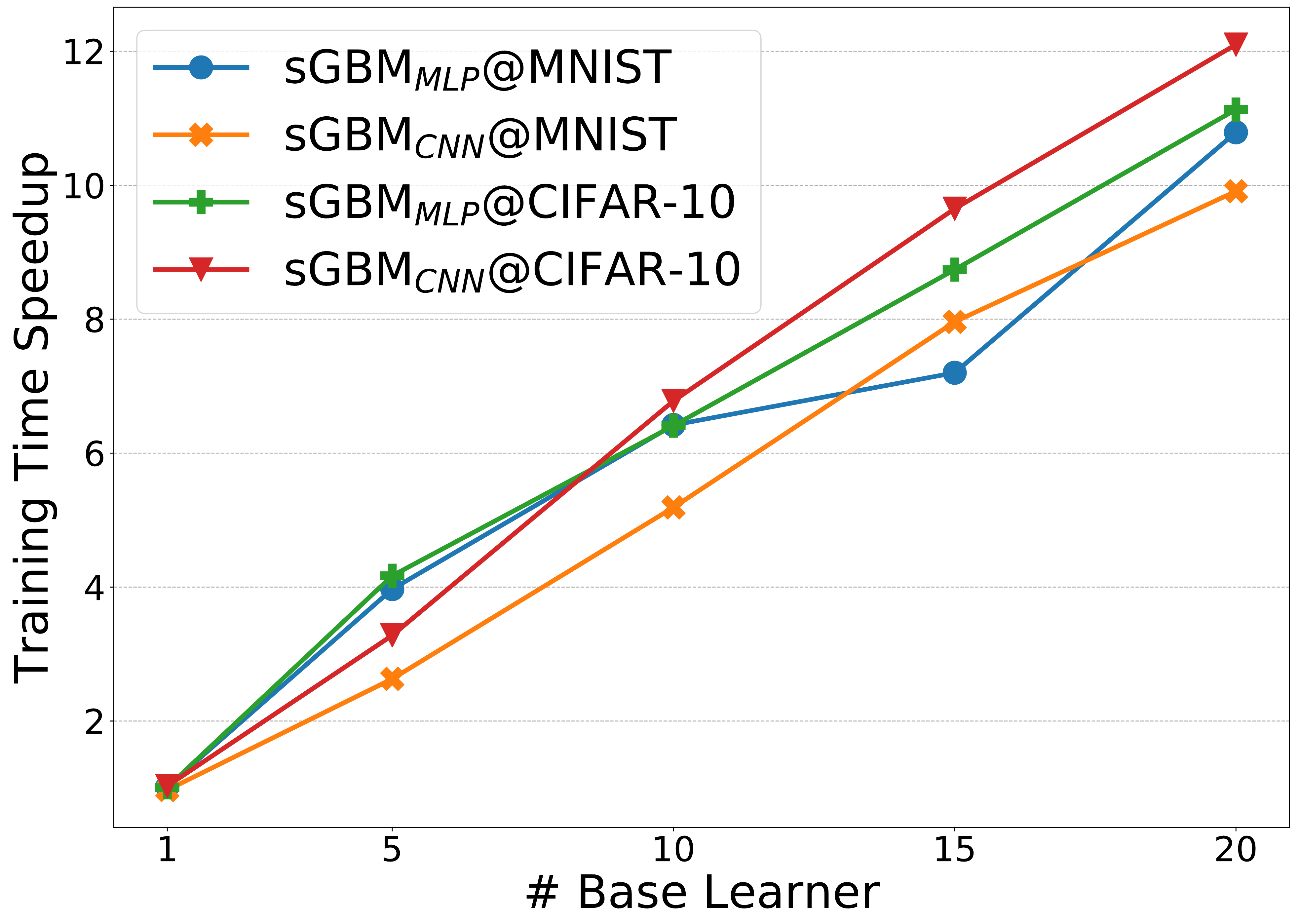}
	\captionof{figure}{Training time speedup}
\end{minipage}	

\begin{table*}[h]
	\centering
	\caption{Classification accuracy (mean$\pm$std) comparison with $10$ base learners (over 5 independent trials).\ N/A indicates the particular base learner is inapplicable.}
	\resizebox{\columnwidth}{!}{%
	\begin{tabular}{rrrrrrr}
		\toprule
		\multicolumn{1}{r|}{} & \emph{Yeast} & \emph{Letter} & \emph{USPS} & \emph{MNIST} & \emph{Fashion-MNIST} & \emph{CIFAR-10} \\ \midrule
		\multicolumn{1}{r|}{GBM$_{MLP}$} & 61.03$\pm$0.56 & 95.87$\pm$0.17 & 96.09$\pm$0.10 & 98.53$\pm$0.05 & 90.61$\pm$0.17 & 55.89$\pm$0.42 \\
		\multicolumn{1}{r|}{sAveraging$_{MLP}$} & 59.73$\pm$0.68 & \textbf{96.13$\pm$0.14} & 95.11$\pm$0.07 & 98.72$\pm$0.07 & 90.21$\pm$0.09 & \textbf{58.33$\pm$0.15} \\
		\multicolumn{1}{r|}{sGBM$_{MLP}$} & \textbf{61.48$\pm$0.19} & 95.86$\pm$0.21 & \textbf{96.30$\pm$0.15} &  \textbf{98.73$\pm$0.04} &  \textbf{90.75$\pm$0.13} & 57.31$\pm$0.26 \\ \midrule
		\multicolumn{1}{r|}{XGBoost} &58.57$\pm$1.12 & 90.61$\pm$0.39 & \textbf{94.80$\pm$0.37} & 94.61$\pm$0.25 & 86.42$\pm$0.19 & 42.80$\pm$0.50 \\
		\multicolumn{1}{r|}{sAveraging$_{Tree}$} & 61.26$\pm$0.43 & \textbf{94.50$\pm$0.11} & 93.97$\pm$0.10 & 95.78$\pm$0.10 & 86.56$\pm$0.18 & 47.12$\pm$0.22 \\
		\multicolumn{1}{r|}{sGBDT} & \textbf{62.02$\pm$0.83} & 88.83$\pm$0.24 & 94.74$\pm$0.19 & \textbf{97.18$\pm$0.16} & \textbf{88.47$\pm$0.05} & \textbf{51.35$\pm$0.33} \\ \midrule
		\multicolumn{1}{r|}{GBM$_{CNN}$} & N/A & N/A & N/A & 99.40$\pm$0.06 & 92.43$\pm$0.24 & 77.78$\pm$0.12 \\
		\multicolumn{1}{r|}{sAveraging$_{CNN}$} & N/A & N/A & N/A & 99.50$\pm$0.02 & \textbf{92.99$\pm$0.09} & 76.97$\pm$0.61 \\
		\multicolumn{1}{r|}{sGBM$_{CNN}$} & N/A & N/A & N/A & \textbf{99.55$\pm$0.02} & 92.85$\pm$0.11 & \textbf{79.70$\pm$0.24} \\ \bottomrule
	\end{tabular}
	}
\label{GBM and sGBM}
\end{table*}

The experimental results can be found in Table \ref{GBM and sGBM}. It can be shown that sGBM$_{CNN}$ outperforms GBM$_{CNN}$ on image classification tasks, and sGBM$_{MLP}$ outperforms GBM$_{MLP}$ on almost all data sets except Letter dataset. For trees, sGBDT only achieved sub-optimal solution on Letter and USPS datasets, compared with the classical XGBoost model.  

\begin{figure}[t]
		\centering
		\includegraphics[width=\columnwidth]{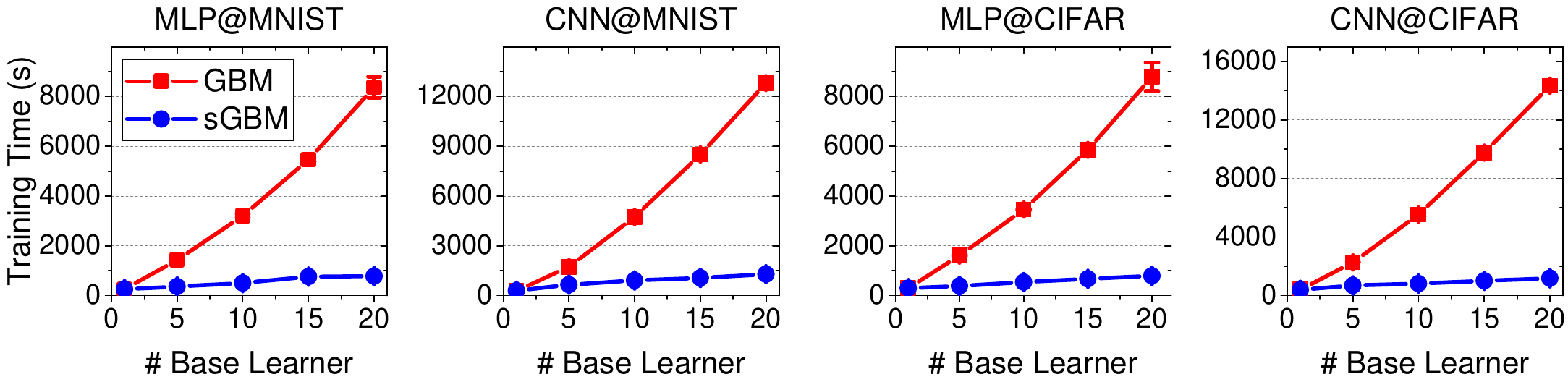}
		\caption{Training time (in seconds) using MLP and CNN as base learners}
		\label{training time}
\end{figure}

To further examine the training time efficiency, we measured the training time of GBM$_{MLP}$ and GBM$_{CNN}$ on MNIST and CIFAR data set by varying the number of base learners during training. Experimental results are summarized in Figure 3 and Figure \ref{training time}. It can be shown that sGBM achieved nearly linear speed up, compared with hard gradient boosting machines when using same base learners. Actually, such result is not so surprising since instead of training base learners one at a time, sGBM naturally train all the base learners simultaneously, a huge gain in terms of training time reduction. 

\begin{figure*}[t]
	\centering
	\includegraphics[width=\columnwidth]{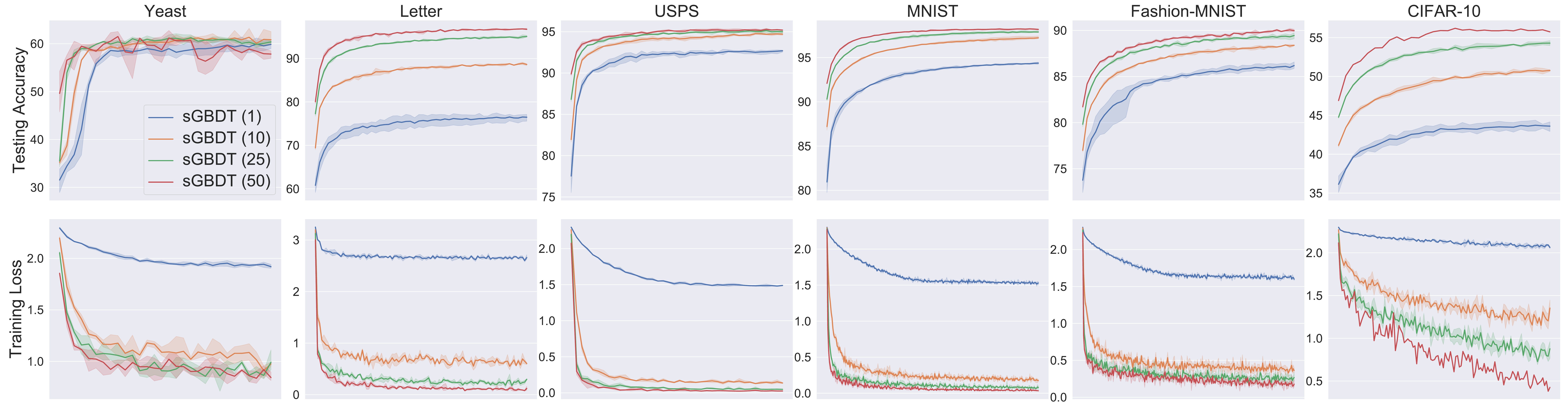}
	\caption{Training curves of sGBDT with adding more base trees. sGBDT($N$) means that $N$ base trees were used (over $5$ independent trials).}
	\label{sGBDT Learning Curve}
\end{figure*}

\subsection{Does adding more base learners help?}

According to the architectural design of sGBM, interactions between base learners were made on purpose, making it interesting to investigate what is the contribution per individual base learners? In other words, how much performance gain can we have when adding more base learners into the system?

To examine this, we plotted the training and testing curves of sGBDT using different number of trees, as illustrated in Figure \ref{sGBDT Learning Curve}. It can be shown that, when adding more base learners into the system, a performance gain during testing time is always observed. In addition, more complex tasks such as CIFAR-10 data, a clearer separation for both training and testing are at presence.  

\subsection{Multi-Output Regression with sGBDT}

Current implementations of hard GBDT including XGBoost cannot handle multi-output regression task efficiently. This is mainly due to the fact that each tree only handles one dimension only. When facing multi-output regression tasks using hard GBDTs such as XGBoost, one possible modification is to treat each target variable independently and train several GBDTs accordingly. Soft GBDT, on the other hand, can be naturally used in such settings without extra modifications.

To compare the performance, we evaluated them on several benchmark datasets\footnote{\texttt{mulan.sourceforge.net/datasets-mtr.html}}, using mean squared error on test sets as evaluation metrics, and the results are presented in Table \ref{mo_exp}. Both XGBoost-MO (multi-output extension) and sGBDT used 10 trees with depth 5 for all datasets. For XGBoost-MO, the learning rate was chosen from $\{0.1, 0.3, 0.5, 1\}$ with best performance. For sGBDT, we used Adam optimizer \cite{kingma2014adam} with learning rate as $10^{-3}$ and weight decay as $5\times10^{-4}$.

\begin{table}[h]
	\centering
	\caption{Mean squared error (mean$\pm$std) with 10 base learners (over 5 independent trials)}
    \begin{tabular}{ccc} \toprule
        \multicolumn{1}{r|}{Dataset} & sGBDT  & XGBoost-MO \\ \midrule
        \multicolumn{1}{r|}{\emph{scm1d}}   & \textbf{.0981 $\pm$ .0014} & .1302 $\pm$ .0045     \\
        \multicolumn{1}{r|}{\emph{scm20d}}  & \textbf{.1371 $\pm$ .0019} & .2595 $\pm$ .0042     \\
        \multicolumn{1}{r|}{\emph{wq}}      & \textbf{.8608 $\pm$ .0033} & .8874 $\pm$ .0279     \\
        \multicolumn{1}{r|}{\emph{enb}}     & \textbf{.0082 $\pm$ .0008} & .0116 $\pm$ .0018     \\
        \multicolumn{1}{r|}{\emph{oes10}}   & .3671  $\pm$ .0167 & \textbf{.2784 $\pm$ .1010}    \\
        \bottomrule
    \end{tabular}
    \label{mo_exp}
\end{table}

Experimental results are summarized in Table \ref{mo_exp}, it can be shown that sGBDT achieves better performance on multi-output regression on most datasets. Notice that sGBDT can be naturally plugged into such tasks with ease whereas the hard GBDT requires extra modifications. 

\subsection{Incremental Learning with sGBDT}

\begin{figure}[ht]
	\centering
	\includegraphics[width=\columnwidth]{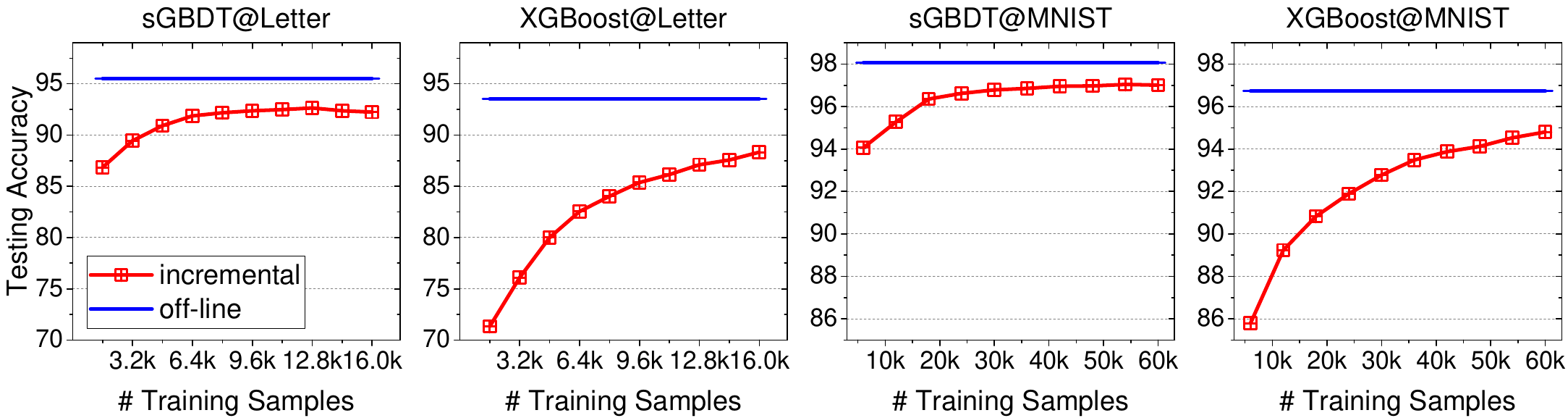}
	\caption{Performance comparison under incremental learning setting}
	\label{incremental_experiment}
\end{figure}

In real world situations, data often arrive with batches, therefore, how the model adapt with the environment is a practical challenge. In this section, we simulated this situation using Letter and MNIST dataset as follows: we divide the training data into 10 equal sized batches and updating the corresponding models using the data received so far and evaluate it on the test set. Both models we used used the same hyper-parameters as in previous sections. We then compare its performance against one model trained on the whole training data (i.e., off-line). The experimental results are summarized in Figure \ref{incremental_experiment}. sGBDT has a clear advantage over XGBoost in terms of a much faster convergence rate. Furthermore, smaller accuracy reduction can be observed for sGBDT, compared with using XGBoost.

\subsection{Knowledge distillation with sGBDT}

It is well accepted that tree-like model is not the best fit for image data, compared with convolutional neural networks. In this section, we test the knowledge distillation ability \cite{zhou2004nec4,hinton2015distilling} of sGBDT models. Concretely, given some trained conv-nets, we are interested in how much knowledge can a much simpler sGBDT with only 10 trees extract from that trained model, using the twice labelling technique for distillation. 

To do so, we first trained an ensemble of LeNet-5 on MNIST and one reduced version of ResNet-18 on CIFAR-10 as the teacher model. Then, as introduced in \cite{hinton2015distilling}, we let the teacher model to re-label the training set with temperature 20 and 1, respectively, and train the sGBDT on the new training set. We also trained the another sGBDT with same model complexity on the original training set. For a comparison, we used XGBoost modified for multi-output regression (since the new labels are class distribution instead one hot classification signals) for knowledge distillation. Results are summarized in Figure \ref{KD Results}. 

\begin{table}[h]
	\centering
	\caption {Knowledge distillation over 5 independent trials. Performance measured as test accuracy.}
	\begin{tabular}{rrr}
		\toprule
		\multicolumn{1}{r|}{} & \emph{MNIST} & \emph{CIFAR-10}  \\ \midrule
		\multicolumn{1}{r|}{Teacher Model} & 99.45 & 94.86 \\ \midrule
		\multicolumn{1}{r|}{sGBDT without KD} & 97.18 $\pm$ 0.16 & 51.35 $\pm$ 0.33 \\
		\multicolumn{1}{r|}{sGBDT with KD} & 97.57 $\pm$ 0.17 & 52.50 $\pm$ 0.23 \\ \midrule
		\multicolumn{1}{r|}{XGBoost without KD} & 94.61 $\pm$ 0.25 & 42.80 $\pm$ 0.50 \\
		\multicolumn{1}{r|}{XGBoost with KD} & 91.94 $\pm$ 0.25 & 42.74 $\pm$ 0.47 \\\bottomrule
	\end{tabular}
	\label{KD Results}
\end{table}

From the experimental results, it can be shown that sGBDT can indeed distill extra knowledge from the trained CNNs. On the other hand, XGBoost cannot enjoy such distillation, on MNIST dataset, the performance even dropped when using the re-labeled dataset. We believe this is because XGBoost or other GBDT implementations used hard CART tree as base models, and when doing multi-dimensional regression tasks, there is less interactions among trees which responsible for their targeting dimensions, making it hard to distill the information reside within the label distribution vector. 

\section{Conclusion}

In this paper, we proposed the soft gradient boosting machine (sGBM) by wiring base learners together which can be simultaneously trained. By introducing local and global objectives, such device is capable of doing gradient boosting in a much faster fashion. Experimental results showed that, sGBM has several advantages in terms of training time efficiency, model accuracy, online learning and knowledge distillation ability. Its variant, the soft gradient boosting trees (sGBDT) can be regarded as an alternative version of XGBoost or its variant models. Theoretical analysis for this new device is can be build upon for future studies.

\bibliography{reference}
\bibliographystyle{alpha}
\end{document}